\documentclass{article}
\usepackage{spconf,amsmath,graphicx,amsfonts,booktabs,amssymb}

\title{Physics Encoded Spatial and Temporal Generative Adversarial Network for Tropical Cyclone Image Super-resolution}
\name{Ruoyi Zhang$^{1}$\thanks{$^{\dagger}$Corresponding author.}, Jiawei Yuan$^{1}$, Lujia Ye$^{1,2}$, Runling Yu$^{3}$, Liling Zhao$^{1,\dagger}$}

\address{$^1$Nanjing University of Information Science and Technology.\\$^2$University of Reading.\\$^3$Shanghai Typhoon Institute, China Meteorological Administration.}
\begin{document}
\ninept
\maketitle
\begin{abstract}

High-resolution satellite imagery is indispensable for tracking the genesis, intensification, and trajectory of tropical cyclones (TCs). However, existing deep learning-based super-resolution (SR) methods often treat satellite image sequences as generic videos, neglecting the underlying atmospheric physical laws governing cloud motion. To address this, we propose a Physics Encoded Spatial and Temporal Generative Adversarial Network (PESTGAN) for TC image super-resolution. Specifically, we design a disentangled generator architecture incorporating a PhyCell module, which approximates the vorticity equation via constrained convolutions and encodes the resulting approximate physical dynamics as implicit latent representations to separate physical dynamics from visual textures. Furthermore, a dual-discriminator framework is introduced, employing a temporal discriminator to enforce motion consistency alongside spatial realism. Experiments on the Digital Typhoon dataset for 4$\times$ upscaling demonstrate that PESTGAN establishes a better performance in structural fidelity and perceptual quality. While maintaining competitive pixel-wise accuracy compared to existing approaches, our method significantly excels in reconstructing meteorologically plausible cloud structures with superior physical fidelity.

\end{abstract}
\section{Introduction}

Tropical cyclones (TCs) are among the most destructive natural disasters globally\cite{Knapp2010THE}, characterized by intense low-pressure vortices and complex convective systems. Accurate monitoring of their fine-grained structures—such as the eye wall and spiral rainbands—is critical for disaster mitigation and meteorological analysis. While geostationary satellites provide continuous observation, hardware constraints often limit the spatial and temporal resolution of the acquired imagery, hindering the precise analysis of rapid TC evolution. Consequently, enhancing the resolution of satellite cloud images via software algorithms, specifically Super-Resolution (SR), has become a pivotal research direction\cite{8770258}.

In recent years, deep learning has revolutionized image SR. Generative Adversarial Networks (GANs), such as SRGAN\cite{ledig2017photorealisticsingleimagesuperresolution}, have established new benchmarks in recovering realistic textures. However, applying these generic SR methods directly to TC imagery presents significant challenges. Unlike static scenes or rigid body motions in standard video datasets, TC cloud systems exhibit fluid properties governed by complex atmospheric dynamics, including rotation, divergence, and deformation\cite{wang2020physics}. Purely data-driven approaches tend to generate several bad cases such as artifacts that look visually plausible but violate physical laws (discontinuous cloud flow or physically impossible deformation), thereby reducing their reference value for meteorologists.

To bridge the gap between data-driven learning and physical principles, Physics-Informed Neural Networks (PINNs)\cite{raissi2019physics} have emerged as a promising paradigm. This trend has rapidly evolved from Physics-Guided Neural Networks (PGNNs)\cite{daw2022physicsguidedneuralnetworkspgnn}, which typically impose physical consistency via soft constraints from loss functions, to the deeper integration found in Physics-Encoded Neural Networks (PENNs)\cite{debezenac2018deeplearningphysicalprocesses,guen2020disentanglingphysicaldynamicsunknown}, where physical principles are explicitly or implicitly embedded into the network architecture itself. These methods have been proved useful in the past years. 

In this paper, we propose the Physics Encoded Spatial and Temporal Generative Adversarial Network (PESTGAN). This framework is designed to address the challenges of physical inconsistency and temporal flickering in satellite imagery. Adopting the PENN philosophy, we construct a Physics Encoded Generator (PEG) with a disentangled architecture\cite{denton2017unsupervised}. This generator separates the modeling of physical dynamics from visual texture synthesis. At its core, we introduce the PhyCell module\cite{guen2020disentanglingphysicaldynamicsunknown}, which mathematically approximates the atmospheric vorticity equation by mapping partial differential operators to constrained convolution kernels. This guides the network to predict cloud movements that align with fluid dynamics. Furthermore, to ensure temporal coherence, we incorporate a dual-discriminator framework featuring a temporal discriminator that penalizes motion inconsistencies between consecutive frames.

The main contributions of this work can be summarized as follows:

\begin{itemize}
    \item \textbf{Physics-Encoded architecture:} We propose a disentangled generator that integrates a PhyCell module, where constrained convolutional kernels simulate differential operators and the resulting approximate physical dynamics are encoded as implicit latent representations to guide super-resolution with atmospheric physical priors.
    \item \textbf{Spatio-Temporal adversarial learning:} Follow previous works\cite{Chu_2017,Chu_2020}, we design a dual-discriminator framework (spatial discriminator and temporal discriminator), enabling the model to generate high-resolution images that are both spatially sharp and temporally coherent.
    \item \textbf{Hybrid objective function:} We formulate a composite loss function that synergizes adversarial loss, reconstruction loss, and physical related loss. This hybrid objective guides network optimization to achieve a balance between perceptual realism, pixel-wise fidelity, and meteorological validity.
\end{itemize}

\label{sec:intro}

\begin{figure*}[htbp]
    \centering
    \includegraphics[width=0.8\linewidth]{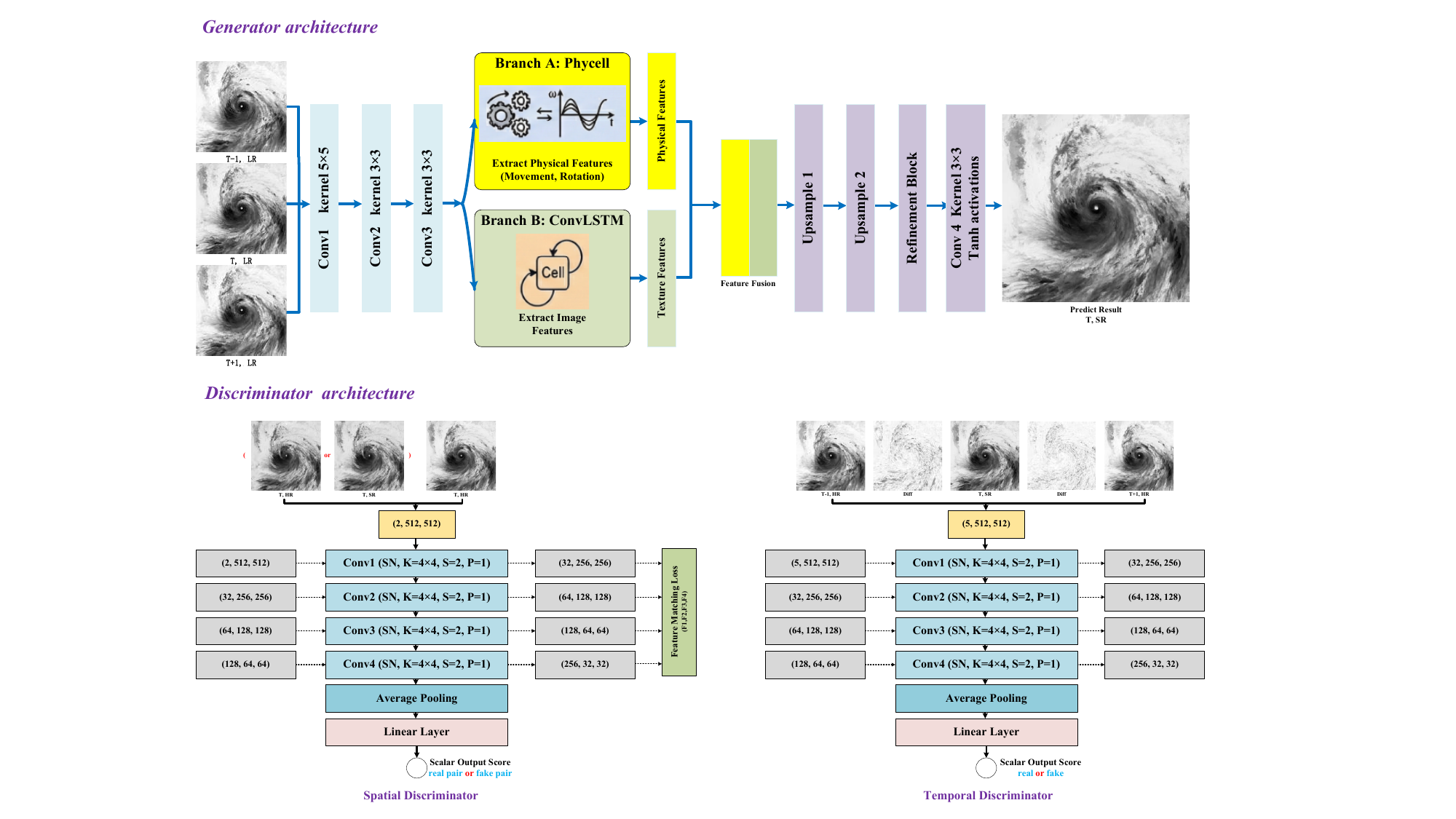}
    \caption{Overall architecture of PESTGAN, the LR pictures are first upsampled and then sent into the physics-encoded generator. After generation, the SR pictures will be sent into a dual-discriminator framework, which is designed for spatial fidelity and temporal coherence.}
    \vspace{-12pt}
    \label{fig:Structure}
\end{figure*}

\vspace{-8pt}
\section{Methodology}

\vspace{-8pt}
\subsection{Overall architecture}
\vspace{-4pt}
The proposed PESTGAN aims to super-resolve a low-resolution (LR) satellite image sequence $\mathcal{I}^{LR} = \{I_{t-1}^{LR}, I_t^{LR}, I_{t+1}^{LR}\}$ into a high-resolution (HR) image $I_t^{SR}$. As shown in Figure \ref{fig:Structure}, the model consists of a Physics Encoded Generator (PEG) and a Dual-Discriminator system. The generator employs a disentangled learning strategy to separate physical flow from visual textures. The training is supervised by a Spatial Discriminator ($D_S$) ensuring perceptual quality and a Temporal Discriminator ($D_T$) enforcing motion consistency. Unlike other models that require pre-computed 3D fluid velocities, our method intrinsically learns 2D latent physical dynamics via the PhyCell module, significantly reducing the cost of training and inference.

\vspace{-8pt}
\subsection{Physics encoded generator}
\vspace{-6pt}
As is known, a promising research direction is to leverage physical knowledge to improve deep learning models. In order to better supervise the generator to produce more realistic and meteorologically-application-friendly typhoon super-resolution results, the core innovation of PEG is the Disentangled Dual-Branch Architecture. We assume the latent representation of a TC cloud system $H$ can be decoupled into physical dynamics $h_{phy}$ and residual textures $h_{res}$.

First, the input low-resolution (LR) frames are explicitly upsampled to the target resolution via nearest-neighbor interpolation before being sent into the first convolutional layer. This step establishes a high-dimensional spatial grid, providing a dense structural prior for the subsequent network. These upsampled frames are then processed by a shared encoder, which progressively maps the visual data into a compact latent feature space through a series of strided convolutions. This encodes the high-dimensional inputs into manageable feature representations that are subsequently fed into two parallel branches.

\vspace{-10pt}
\subsubsection{Disentangled architecture}

\textbf{Branch A (Physical dynamics):} This branch is responsible for capturing the macro-scale motion and deformation of the cloud system, which follows atmospheric laws. We employ the PhyCell module, a recurrent unit designed to model partial differential equations (PDEs) in the latent space. The module utilizes larger kernels ($7\times7$) to capture long-range spatial dependencies inherent in fluid dynamics. It outputs the physical state $h_{phy}$ at the center frame. 

\textbf{Branch B (Residual texture):} This branch focuses on generating high-frequency local details and correcting non-physical visual artifacts. We utilize a convolutional LSTMCell\cite{shi2015convolutionallstmnetworkmachine} with smaller kernels ($3\times3$) to capture local pixel variations. Unlike the physical branch, the residual branch is unconstrained, allowing it to learn the complex, chaotic texture patterns of TCs that are difficult to model with simplified PDEs.

Finally, the outputs from both branches at the target time step $t$ are fused. We concatenate $h_{phy, t}$ and $h_{res, t}$ and pass them through a fusion convolution layer. A decoder consisting of transposed convolution layers and residual blocks then reconstructs the final super-resolved image $I^{SR}_t$, ensuring the result possesses both physically correct motion trends and realistic cloud textures.

\vspace{-8pt}
\subsubsection{Phycell injection}

To inject physical inductive bias into Branch A without relying on explicit physical-field inputs, we integrate the PhyCell module originally proposed in PhyDNet\cite{guen2020disentanglingphysicaldynamicsunknown}. Unlike standard recurrent units that learn black-box temporal transitions, PhyCell performs prediction--correction in the latent space using convolutional operators that are constrained to behave like partial differential operators (PDE operators)\cite{long2018pdenetlearningpdesdata}.

In our PESTGAN, PhyCell should be viewed as an \emph{operator-learning} module rather than a simulator that requires observed physical quantities. Specifically, we impose moment constraints on its internal convolutional kernels so that each kernel approximates a target derivative operator ($\frac{\partial}{\partial x}$, $\frac{\partial^2}{\partial x^2}$, ...) under the well-known correspondence between differentiation and convolution\cite{dong2017image,cai2012image}. By composing these constrained kernels, PhyCell can represent a rich family of dynamical operators and learn their linear combinations directly from image sequences.

In particular, we leverage the tropical-cyclone vorticity equation as the physical prototype to be \emph{approximated} in latent space:
\begin{equation}
\frac{D\zeta }{Dt}=-(\zeta +f)\nabla \cdot \mathbf{v}+\frac{1}{{{\rho }^{2}}}(\nabla \rho \times \nabla p)\cdot \mathbf{k}+\nu {{\nabla }^{2}}\zeta 
\label{eq:phy}
\end{equation}
where $\mathbf{v}, \rho, p, f, \nu$, and $\mathbf{k}$ denote velocity vector, density, pressure, Coriolis parameter, kinematic viscosity, and vertical unit vector, respectively.

Importantly, our implementation does \emph{not} require explicit physical variables as inputs, nor does it compute each term in Eq.~(\ref{eq:phy}) from such measurements. Instead, by regularizing the kernels with $\mathcal{L}_{ker}$, PhyCell learns a set of \emph{composable differential operators} in latent space, and applies them recurrently to produce an \emph{approximate physical prediction} ( $\tilde{h}_{t+1}$) that serves as an implicit physical latent encoding. This encoded approximation then guides the generator toward dynamics-aware reconstruction rather than pure texture hallucination.

\begin{figure}[htbp]
    \centering
    \includegraphics[width=1\linewidth]{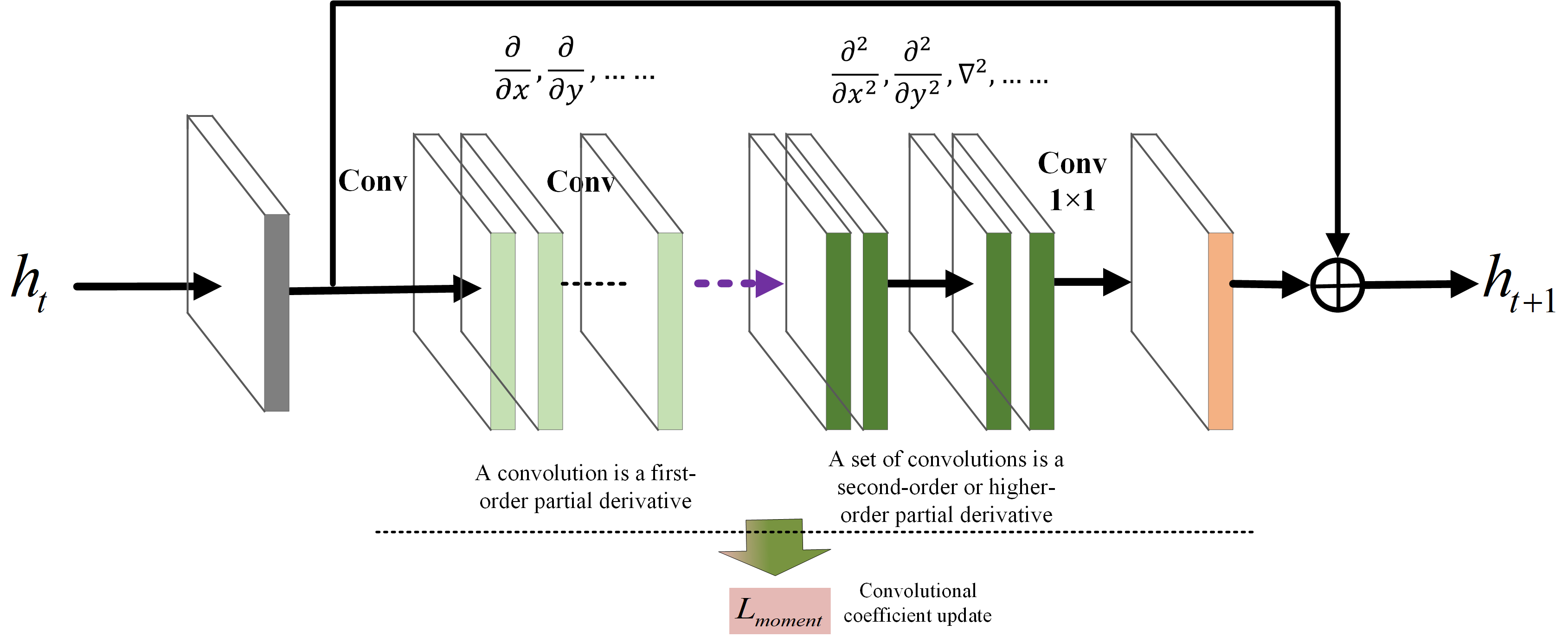}
    \caption{The structure of PhyCell}
    \vspace{-4pt}
    \label{fig:phycell}
\end{figure}

In Figure \ref{fig:phycell}, $h_t$ denotes the latent state at time $t$, and $x_t$ denotes the current input feature (encoded from the upsampled LR frame). PhyCell follows a prediction--correction mechanism.

In the \emph{prediction} step, the physics-encoded operators produce an intermediate state $\tilde{h}_{t+1}$ by applying constrained convolutions that approximate partial derivatives and their compositions. This predicted state can be interpreted as the encoded result of the approximated physical dynamics.

In the \emph{correction} step, the predicted state is adjusted using the current observation $x_t$ via a learnable gain $K$:
\begin{equation}
    h_{t+1} = \tilde{h}_{t+1} + K\,(x_t-\tilde{h}_{t+1}).
\end{equation}
During training, $\mathcal{L}_{ker}$ regularizes the convolutional filter coefficients so that they implement the intended operator-like behavior.

\vspace{-6pt}
\subsection{Dual discriminators}
\vspace{-6pt}
To ensure that the generated super-resolved sequences are not only realistically textured but also temporally consistent with atmospheric fluid dynamics, we employ a dual-discriminator framework. This framework consists of a Spatial Discriminator ($D_S$) and a Temporal Discriminator ($D_T$), both optimized using the Hinge loss to stabilize adversarial training\cite{brock2019largescalegantraining}.

\subsubsection{Spatial discriminator}

The Spatial Discriminator $D_S$ is designed to distinguish between real high-resolution satellite images and the generated frames, focusing on recovering high-frequency spatial details.
As Figure \ref{fig:Structure} shown, $D_S$ adopts a fully convolutional architecture. It takes the concatenation of the upsampled LR frame and the candidate HR frame (real or generated) as input. To stabilize the training process and satisfy the Lipschitz continuity constraint, we apply spectral normalization (SN)\cite{miyato2018spectralnormalizationgenerativeadversarial} to the convolutional layers, except for the final layer.
The network consists of a series of strided convolution blocks that progressively downsample the feature maps, followed by a final linear layer to output a scalar realism score. We empirically find that, under the Hinge loss, removing SN from the last layer yields better performance. Furthermore, $D_S$ is also tasked with extracting intermediate feature maps to compute the Feature Matching Loss\cite{wang2018highresolutionimagesynthesissemantic}, which forces the generator to align with the perceptual representation of real TC imagery.

\subsubsection{Temporal discriminator}

A critical challenge in video super-resolution is preventing temporal flickering and ensuring that cloud movements adhere to physical continuity. Standard video discriminators often rely on 3D convolutions to implicitly learn motion patterns, which can be computationally expensive and difficult to converge.
Instead, we design a lightweight yet effective Temporal Discriminator $D_T$ that explicitly leverages motion cues.
Specifically, for a sequence of three consecutive frames $\{I^{HR}_{t-1}, I^{SR}_t, I^{HR}_{t+1}\}$ where $I_t$ can be either real or fake, we explicitly calculate the temporal finite differences to capture instantaneous motion dynamics:
$$
\Delta^{SR}_{prev} = I^{SR}_t - I^{HR}_{t-1}, \quad \Delta^{SR}_{next} = I^{HR}_{t+1} - I^{SR}_t
$$
The input to $D_T$ is constructed by concatenating the raw frames and their corresponding difference maps along the channel dimension, resulting in a 5-channel tensor $\{I^{HR}_{t-1}, I^{SR}_t, I^{HR}_{t+1}, \Delta^{SR}_{prev}, \Delta^{SR}_{next}\}$ (since the Digital Typhoon Dataset's images are single channel). This design forces the discriminator to scrutinize both the visual content and the frame-to-frame residuals. If the generated sequence exhibits unnatural jittering or discontinuous flow that violates the continuity equation, the difference maps will amplify these artifacts, allowing $D_T$ to effectively penalize physically inconsistent motion. Similar to $D_S$, Spectral Normalization is applied to ensure training stability.

\vspace{-8pt}
\subsection{Loss Function}
\vspace{-6pt}
\label{sec:loss}

The network is optimized via a hybrid objective function composed of three categories: reconstruction constraints, adversarial learning, and physics-encoded regularization. This composite ensures a balance between pixel-wise fidelity, perceptual realism, and meteorological validity. The total loss $\mathcal{L}_{total}$ is formulated as:

\begin{equation}
\begin{aligned}
\mathcal{L}_{total} = \;& \lambda_{1}\mathcal{L}_{1} + \lambda_{feat}\mathcal{L}_{feat} + \lambda_{adv}\mathcal{L}_{adv} \\
& + \lambda_{stat}\mathcal{L}_{stat} + \lambda_{ker}\mathcal{L}_{ker}
\end{aligned}
\label{eq:total_loss}
\end{equation}

\noindent where $\lambda$ terms are hyperparameters balancing the contribution of each component.

\textbf{Reconstruction Constraints.} 
To ensure basic structural consistency, we employ the pixel-wise $L_1$ loss, denoted as $\mathcal{L}_{1} = ||I^{SR} - I^{HR}||_1$. However, relying solely on pixel distance often leads to overly smooth textures. To mitigate this, we incorporate a Feature Matching Loss ($\mathcal{L}_{feat}$)\cite{wang2018highresolutionimagesynthesissemantic}, which minimizes the Euclidean distance between the intermediate feature maps of the discriminator for real and generated sequences. This forces the generator to capture high-level structural patterns that are perceptually significant.

\textbf{Adversarial Learning.} 
To recover realistic cloud textures and coherent motion, we adopt a dual-discriminator scheme. The adversarial loss $\mathcal{L}_{adv}$ is the sum of spatial and temporal components based on the Hinge loss variant. The spatial discriminator $D_S$ penalizes unrealistic static textures, while the temporal discriminator $D_T$ takes consecutive frames (and their differences) as input to detect motion inconsistencies. The generator minimizes:

\vspace{-8pt}
\begin{equation}
\mathcal{L}_{adv} = - \mathbb{E}[D_S(I^{SR})] - \mathbb{E}[D_T(\{I^{HR}_{t-1}, I^{SR}_t, I^{HR}_{t+1}, \Delta^{SR}_{prev}, \Delta^{SR}_{next}\})]
\end{equation}

This drives the model to generate sharp details and smooth transitions indistinguishable from real satellite observations.

\textbf{Physics-Encoded Constraints.} 
A core contribution of PESTGAN is explicit physical regularization, consisting of two terms inspired by previous works\cite{guen2020disentanglingphysicaldynamicsunknown,long2018pdenetlearningpdesdata}:

\textit{[1] Kernel Moment Loss ($\mathcal{L}_{ker}$):} To let PhyCell \emph{learn} physics-inspired latent dynamics from image sequences, we constrain its convolutional kernels to behave as a basis of partial differential operators, thereby explicitly embedding physical priors into PhyCell. With these operator-like kernels, the recurrent transition can express (and thus approximate) the operator composition implied by Eq.~(\ref{eq:phy}) in latent space, without requiring explicit physical-variable inputs. Following the Moment Analysis theory, a convolution kernel $W$ approximates a differential operator if its moment matrix $M(W)$ matches a target geometric moment pattern. We impose:

\begin{equation}
\mathcal{L}_{ker} = \sum_{k} || M(W_k) - M_{target} ||_F^2
\end{equation}
where $W_k$ represents the filters in the PhyCell, and $M_{target}$ defines the ideal derivative coefficients derived from the finite difference method. This ensures the learned latent dynamics adhere to fluid mechanics.

\textit{[2] Statistical Consistency Loss ($\mathcal{L}_{stat}$):} To ensure meteorological validity, we impose statistical constraints on both texture distribution and motion smoothness:

\vspace{-6pt}
\begin{equation}
     \mathcal{L}_{stat} = \underbrace{\| \sigma^2(I^{SR}_t) - \sigma^2(I^{HR}_t) \|_2^2}_{\text{Spatial Energy Matching}} + \lambda_{t} \underbrace{\sigma^2(I^{SR}_t - I^{SR}_{t-1})}_{\text{Temporal Continuity}}
\end{equation}
\vspace{-8pt}

\noindent where $\sigma^2(\cdot)$ denotes the variance computed over spatial dimensions. The first term enforces the generator to match the spectral energy of real TC cloud systems, preventing overly smooth textures. The second term penalizes the spatial variance of temporal differences, effectively suppressing non-physical high-frequency flickering while preserving coherent cloud motion.

\vspace{-6pt}
\section{Experiments}
\vspace{-6pt}

\subsection{Dataset and implementation details}
\vspace{-4pt}
All models are trained on the Digital Typhoon dataset\cite{kitamoto2023digitaltyphoonlongtermsatellite} samples from the Western North Pacific. Samples with missing timestamps or too short durations were removed during preprocessing. In total, 15000 infrared cloud images with 512×512 size were selected covering the period from 2018 to 2022. To strictly evaluate generalization capabilities on unseen meteorological events, we reserve the 1st and 14th typhoons of 2022 as the test set, while the remaining sequences are used for training. For quantitative evaluation, we employ Peak Signal-to-Noise Ratio (PSNR) and Structural Similarity Index (SSIM)\cite{1284395} to measure pixel-wise reconstruction fidelity. Additionally, we provide qualitative comparisons to visually demonstrate the model's performance in reconstructing realistic cloud textures and maintaining temporal coherence.

\subsection{Comparison study}

To evaluate the effectiveness of PESTGAN, we compare it against five state-of-the-art video super-resolution methods: TDAN\cite{tian2018tdantemporallydeformablealignment},
EDVR\cite{wang2019edvrvideorestorationenhanced}, BasicVSR\cite{chan2021basicvsrsearchessentialcomponents}, RealBasicVSR\cite{chan2021investigatingtradeoffsrealworldvideo}, and RealViformer\cite{zhang2024realviformerinvestigatingattentionrealworld}. The quantitative results on the test set are summarized in Table \ref{tab:comparison}.

As illustrated in the table, PESTGAN achieves the highest SSIM (0.8656), outperforming the second-best method by a notable margin. This performance gain demonstrates that our physics-encoded architecture excels at preserving the complex structural integrity of tropical cyclones. Moreover, PESTGAN also achieves the best PSNR (30.31 dB) among all compared methods, indicating that introducing physics-encoded and perceptual constraints does not sacrifice pixel-wise fidelity on this dataset. Overall, PESTGAN provides a strong balance between pixel-level accuracy and structural realism.

\vspace{-8pt}
\begin{table}[htbp]
    \centering
    \caption{Evaluation metrics of all methods on test set, higher is better for each metric.}
    \label{tab:comparison}
    \vspace{2mm}
    \begin{tabular}{lcc}
        \toprule
        \textbf{Model} & \textbf{PSNR (dB)} $\uparrow$ & \textbf{SSIM} $\uparrow$ \\ 
        \midrule
        TDAN          & 29.59   & 0.8386   \\
        EDVR          & 29.65   & 0.8391   \\
        BasicVSR      & 30.12   & 0.8401   \\
        RealBasicVSR  & 30.03   & 0.8565   \\
        RealViformer  & 30.10  & 0.8572   \\
        \textbf{PESTGAN (Ours)} & \textbf{30.31}  & \textbf{0.8656}   \\
        \bottomrule
    \end{tabular}
    \vspace{-6pt}
\end{table}

\begin{figure*}[htbp]
    \centering
    \includegraphics[width=0.95\linewidth]{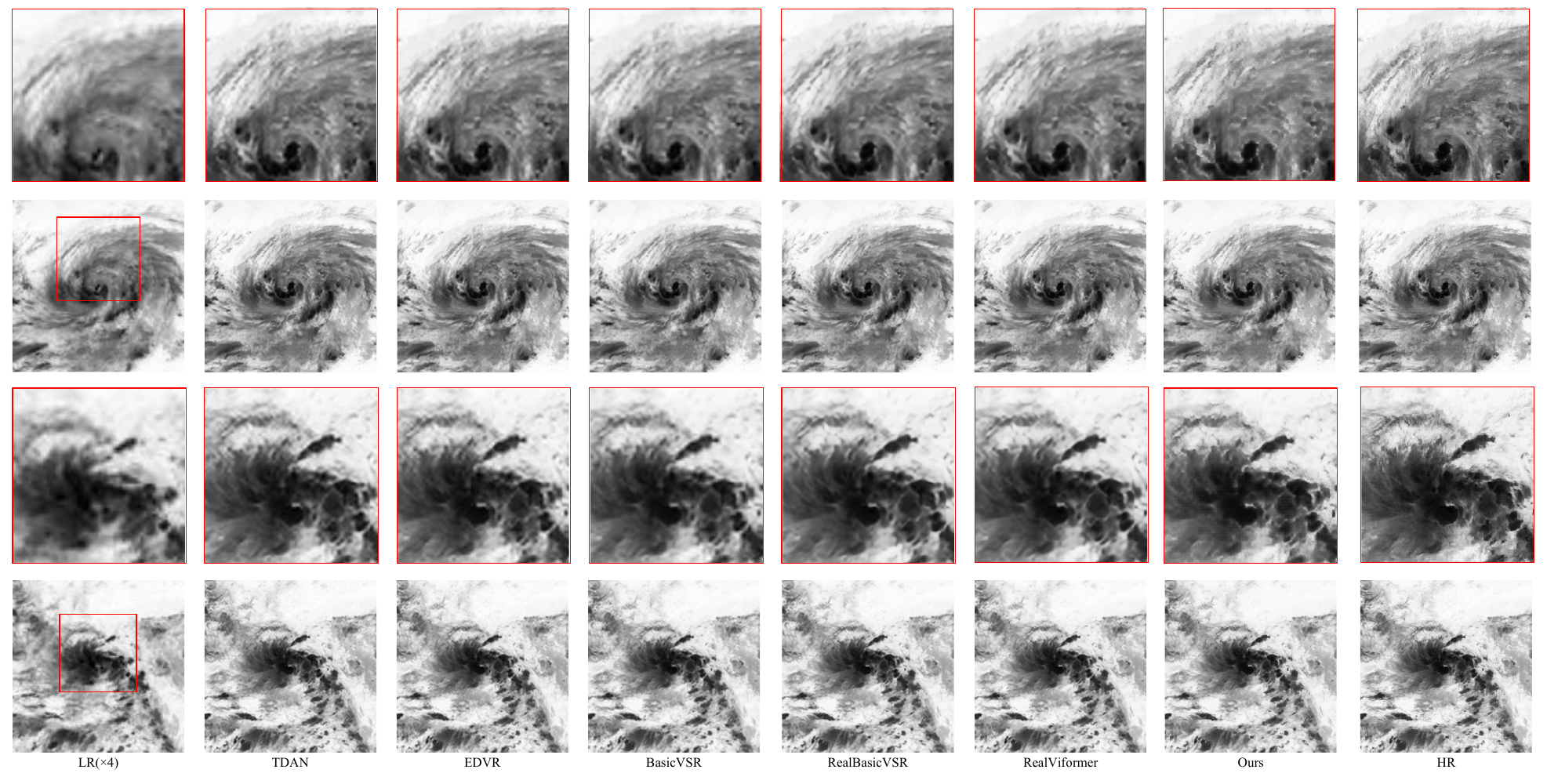}
    \vspace{-8pt}
    \caption{Visual comparison of the super-resolution results of different models on the 4× super-resolution task across various test sequences.}
    \label{fig:compare}
\end{figure*}

Beyond quantitative metrics, we provide a detailed visual comparison in Figure \ref{fig:compare} to assess perceptual quality. As observed in the zoomed-in patches, earlier methods like TDAN and EDVR generate severe mosaic-like patterns. This limitation likely stems from a lack of physical awareness, rendering the networks unable to effectively characterize the complex fluid dynamics in TC imagery. While recent models like RealViformer avoid such artifacts, they exhibit noticeable blurring in high-frequency regions. In contrast, PESTGAN produces significantly clearer results with refined details. This comparison suggests that the high PSNR scores of competing methods may be driven by MSE-dominated optimization, which tends to over-smooth textures rather than truly "understanding" and recovering structural details. These visual results align with our superior SSIM scores, confirming that incorporating physical priors enables PESTGAN to handle these complex meteorological scenarios more effectively.

\vspace{-6pt}
\subsection{Ablation Study}

To measure the contribution of each component to the overall performance, we conducted a series of ablation studies. We also introduced the Relative Average Spectral Error (RASE)\cite{1634730} to evaluate the global reconstruction fidelity, which is an important metric for measuring whether the model preserves meteorologically meaningful structures. The specific metrics for different strategies are presented in Table 2.

\vspace{-8pt}
\begin{table}[htbp]
    \centering
    \caption{Metrics of every model structure, while lower is better for RASE.}
    \label{tab:ablation_minimal}
    \vspace{2mm}
    \begin{tabular}{lccc}
        \toprule
        \textbf{Model} & \textbf{PSNR (dB)} $\uparrow$ & \textbf{SSIM} $\uparrow$ & \textbf{RASE (\%)} $\downarrow$ \\ 
        \midrule
        Baseline           & 29.1662 & 0.8480 & 4.8320 \\
        Model-I            & 28.3924 & 0.8111 & 5.2813 \\
        Model-II           & 29.6262 & 0.8442 & 4.3900 \\
        Model-III          & 28.9266 & 0.8548 & 5.3054 \\
        \textbf{PESTGAN (Ours)} & \textbf{30.3191} & \textbf{0.8656} & \textbf{3.9898} \\
        
        \bottomrule
    \end{tabular}
\end{table}

\begin{figure}
    \centering
    \includegraphics[width=1\linewidth]{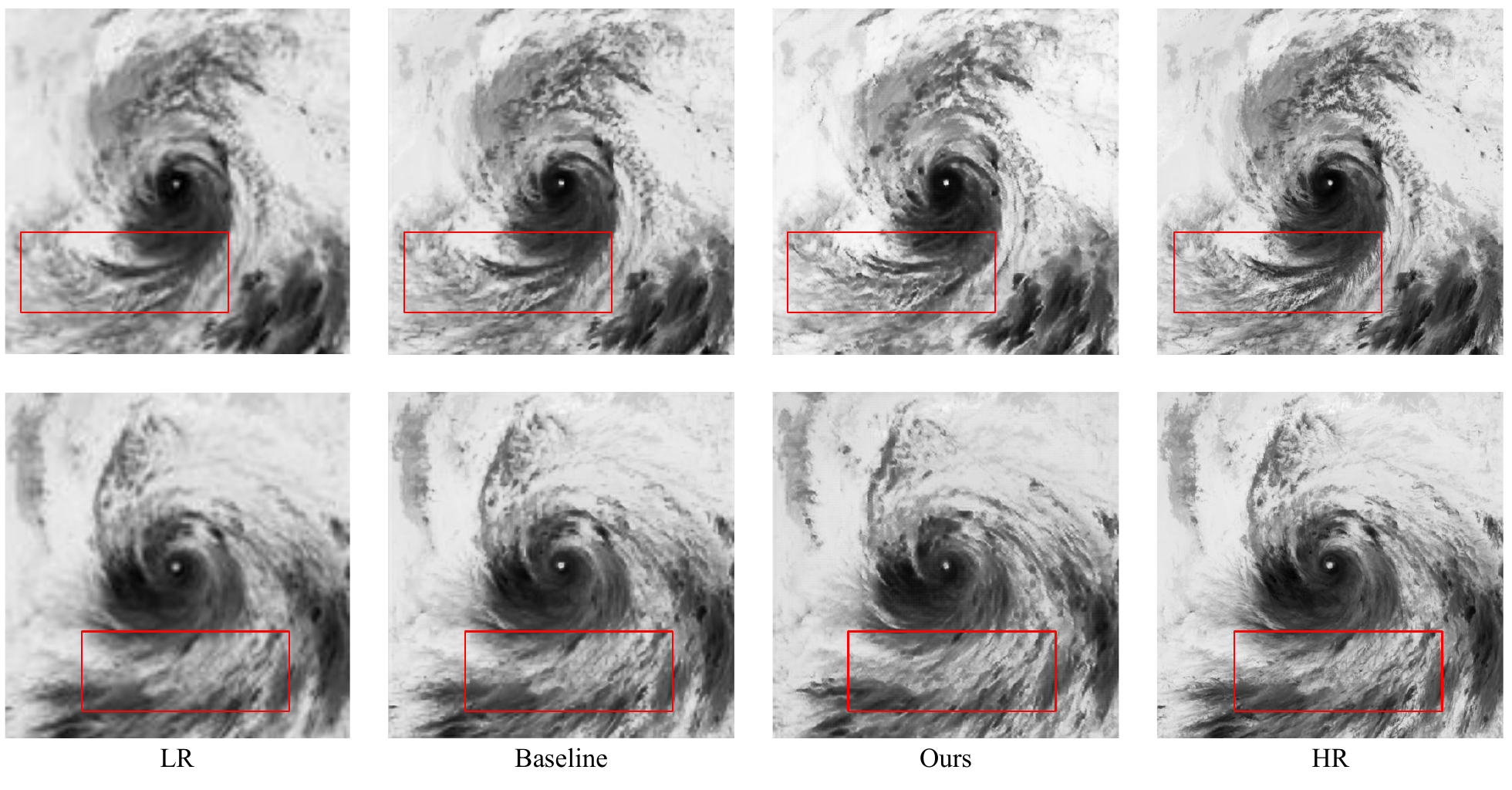}
    \vspace{-10pt}
    \caption{Visual comparison of the super-resolution results of ablation study on the 4× super-resolution.}
    \vspace{-16pt}
    \label{fig:ablation}
\end{figure}

\textbf{Baseline:} We established a baseline using a dual-discriminator GAN similar to TempoGAN\cite{Chu_2017}. It employs a standard ResNet-based generator without internal physical modules. Crucially, it relies on pre-computed physical fields as conditional inputs to guide generation, which incurs high computational costs during inference.

\textbf{Model-I:} We introduced the PhyCell module connected serially with residual blocks. This architecture failed to balance the trade-off between strict physical constraints and visual texture generation, resulting in the worst performance (PSNR=28.39) across all metrics.

\textbf{Model-II:} This variant adopts a physics-guided strategy where a parallel low-resolution PhyCell branch injects dynamic features into the main texture branch. While it improves upon the baseline by incorporating internal physical priors, the lack of explicit disentanglement limits its ability to recover fine details.

\textbf{Model-III:} In this variant, we advanced beyond Model-II by introducing the disentangled dual-branch generator to better separate dynamics from texture. Furthermore, to enforce strict physical consistency, we also integrated the PhyCell module into the discriminator. However, contrary to expectations, performance deteriorated (4.6\% drop in PSNR) compared to PESTGAN. We hypothesize that the high complexity of a physics-encoded discriminator made the optimization landscape overly difficult, preventing the generator from reaching convergence during adversarial training.

\textbf{PESTGAN (Ours):} Derived from the lessons of Model-III, we retained the superior disentangled generator architecture but removed the physical constraints from the discriminator. By pairing the physics-encoded generator with a standard dual-discriminator, PESTGAN achieves the best performance. This confirms that while decoupling physical laws in the generator is crucial for reconstruction quality, enforcing explicit PDEs in the discriminator is unnecessary and potentially destabilizing.

Figure 4 highlights the significant improvement of PESTGAN over the baseline. While the baseline suffers from blurred, blocky artifacts when reconstructing spiral rainbands, our method generates distinct striped structures with varying grayscale intensities that accurately reflect rainband intensity. This superior perceptual and physical fidelity validates the effectiveness of incorporating the proposed physical modules and the hybrid objective function.

\section{Conclusion}
In this paper, we proposed PESTGAN, a physics-encoded framework for tropical cyclone image super-resolution. Our core innovation is a disentangled generator that approximates the vorticity equation via latent-space constrained convolutions, injecting physical inductive bias without requiring explicit meteorological inputs. By synergizing this architecture with a spatio-temporal adversarial strategy, PESTGAN achieves an impressive performance on the Digital Typhoon dataset. The results demonstrate a superior balance between pixel-wise fidelity and meteorological plausibility, successfully recovering complex cloud structures that adhere to underlying fluid dynamics.

\section{Acknowledgements}
This work was supported in part by the Shanghai Typhoon Research Foundation from Shanghai Typhoon Institute of China Meteorological Administration under Grant TFJJ202208, in part by the Innovation andDevelopment Special Program of China Meteorological Administration under Grant CXFZ2024J006.

\bibliographystyle{IEEEbib}
\bibliography{strings,refs}

\end{document}